# Psychological Health Knowledge-Enhanced LLM-based Social Network Crisis Intervention Text Transfer Recognition Method

Shurui Wu *

Weill Cornell Medicine, New York City, NY, USA, shuruiwu215@gmail.com

Xinyi Huang

University of Chicago, Chicago, IL, USA, bellaxinyihuang@gmail.com

Dingxin Lu

Icahn School of Medicine at Mount Sinai, New York, NY, USA, sydneylu1998@gmail.com

**Abstract**

As the prevalence of mental health crises increases on social media platforms, how to effectively identify and deter potential harms has emerged as an urgent problem. To improve the detection ability of crisis-related content in social networks, this study proposes a large language model (LLM) text transfer recognition method for social network crisis intervention based on the enhancement of mental health knowledge that integrates mental health professional knowledge and transfer learning technology. We introduce a multi-level framework that employs transfer learning on a large language model BERT and integrates domain mental health knowledge, sentiment analysis as well as behavior prediction modeling techniques. This approach proposes a mental health annotation tool trained on social media datasets from crisis events, helping a large language model find potential language cues and then determine the presence of a psychological crisis and crisis acts. Experimental results indicate that the proposed model is superior to the traditional method in crisis detection accuracy, and demonstrate a greater sensitivity to underlying differences in context and emotion.

CCS CONCEPTS

• **Applied computing ~ Life and medical sciences** ~ Health care information systems

**Keywords**

Psychological Health Knowledge, Large Language Models (LLMs), Crisis Intervention, Text Transfer Recognition, Transfer Learning

## 1 INTRODUCTION

At present, mental health issues are on the rise worldwide due to the proliferation of social media and internet-based platforms, a phenomenon extensively documented in recent studies on online emotional and psychological support during public health crises [18].Over the last decade the explosive growth of social platforms like Facebook, Twitter, Instagram and TikTok has resulted in increased sharing of personal lives and emotional devastation on those platforms. It has helped many people not just find psychological comfort through social support, but also made mental health issues a public issue. According to the World Health Organization (WHO), mental health disorders have been one of the most important public health problems in the world, especially among young people and social media users, and the increase rate of mental health problems are significantly greater than other groups. Such as mental health problems, including depression, anxiety, suicidal tendency, etc [1]. Overall discussion in the social platform is higher, and the early symptoms are more likely to be recessive and hidden, which also increases the difficulty of timely intervention and effective saving.

Feelings expressed in public, and especially in social networking, are still part of the changing trend of emotion in modern society. On an online social network, users can share their thoughts with any audience from the anonymity of his or her room, while in conventional face-to-face communication, users will be confronted with the expression of his or her counterpart and the tone of voice, etc., the core of which can be ignored, and the psychological distress of the victim. But this anonymity and virtuality is also what makes people so concealed yet complicating their experiences of psychological pain [2]. Many requests for help, crisis signals are camouflaged within confusing textual and nonverbal cues (mood swings, tone changes, rupture and contradiction in self-presentation, and other) That makes it especially challenging for the automation of potential crisis identification.

Expressions in social networks cover broader range of data such as text, images, videos, audio, and other forms of media. Yet text is by far the primary medium of communication throughout all of this, and in textual content including blogs, tweets, comments, and etc. Users can express their emotion, life challenges, and beliefs detailed. However, texts from social networks tend to be noisy and ambiguous. In many of these cases emotions are expressed indirectly, through subtle wording or humor, irony other expressions which makes it hard to gain the polarity, intensity, and shifting of emotions [3]. Furthermore, the emotions that are reflected on social networks by users on the platform are not necessarily consistent with the emotions in reality, and are susceptible to multi-factor influences such as socio-cultural background, personal expression habits and platform use preferences, which increase the difficulty and complexity of emotion recognition and understanding [4].

As a result, social network crisis identification mode has many challenges in existing text analysis technology. There are existing methods for sentiment analysis and crisis recognition which mainly focus on the use of simple keyword matching or basic sentiment classification. These techniques largely miss the richness of emotional traits encoded in language and do not capture the nuances of implicit and complex mental health signals. For instance, some crisis behaviors like depression or suicidal tendencies might appear as slight changes in affect, subtle changes in language, or the user shutting down and avoiding interaction with them [5]. In traditional text analytics approaches, these nuances are often lost, which means we don’t detect some signals of a crisis.

Moreover, despite certain advances in the current crisis intervention system to some degree, most of them were still based on shallow sentiment analysis or keyword-matching methods. And while sentiment analysis techniques can determine the emotional orientation (positive and negative emotions) of text, they generally do not take into account the temporal variation of emotions, contextual differences, and individual variations (situational expressions) in emotional expression. For example, while users pretend to be happy when they are actually sad and inject humor or sarcasm into their expressions, traditional sentiment analysis algorithms very often have a poor accuracy in this case [6]. Existing methods of sentiment analysis are usually far from sufficient when it comes to more complex scenarios like the interlacing of many different emotional states. In judging complex and dynamic emotional and psychological crises in the social network, the existing system cannot be better understood, and there is a lack of sufficient in-depth understanding and situational judgment ability which leads to their effect of intervention can only fit into, the effect is far less than expected [7].

Based on all the above consideration, through the integration of mental health knowledge for the social network text transfer recognition method of crisis intervention, this paper puts forward a transfer recognition method for crisis intervention text based on large language model (LLM). It is based on a combination of domain knowledge in the field of mental health and transfer learning capabilities to improve early detection of a crisis signal in social networks using a multi-level framework. It can not only realize the sensitive mining of emotional fluctuation in social network text, but also through transfer learning technology, transfer the model to the language dynamic environment, to better realize the subtle emotional fluctuation in environmental perception text, language implicit representation, tongue difference in individual social network in the crisis recognition process.

## 2 RELATED WORK

Oktavianus and Lin [8] add to this literature by examining migrant domestic workers who seek social support through social networks during a public health crisis. Studies on the emotional support and coping strategies of temporary migrants in crises such as the pandemic. The study explores how immigrant populations use social media to create emotional support and increase social connections, by analysing storytelling and community interactions across the social media landscape. Indeed, the research points out that, during times of crisis, social media serves to provide mental health support to vulnerable populations, as cyberspace can serve as a cauldron to find belongingness and security.

The study by Bolhuis and Wijk [9] explores the use of social media and mobile devices to assist with asylum processes in five European countries, including the review of immigration applications. The research highlights how migrants and asylum seekers seek to reach the outside world through social media platforms and mobile devices in the digital age, and examines how immigration authorities screen – by checking social media activity and content. In the case of migration management and crisis response, the study draws attention to the critical need for a media tool to be integrated into government life as well as highlight how social media be utilized as a bi-directional information shipper during times of panic or crisis — not only within the public domain of health crises or emergencies, but also in the wider context of social connection and interest.

Furthermore, Lv et al. [10] said that big data has the potential to help in the crisis management of the COVID-19 pandemic. They explain that their study shows the use of social media as a source of big data during the pandemic, which will help with identifying the source of the infection and examining the emotional response surrounding events. The researchers emphasize the importance of social media as an essential tool to shape public health in terms of social media, text mining, predicting analytics, and social network analytics. The study demonstrates that big data technology can be applied in crisis management particularly in global health crises (epidemics) to ensure predictive outcomes of population behaviour, emotional trends and health response.

Jin and Spence analyzed Twitter [11] tweets using the CERC (Crisis and Emergency Risk Communication) model. Using thematic modeling, the study explores social media's dissemination of information as well as its organization of crisis communication and public reactions to crisis management in the wake of a disaster. Through this analysis of tweets surrounding Harvey Maria, it showcases the various affairs of social media users on how people voice their emotions, fears, anger and confusion, all the while noting the different matters that platforms serve during crises, including information dissemination, emotional expression and public emotion management.

Wildemann et al. [12] applied large-scale social media text analytics to discover movement changes by applying sentiment analysis techniques to unveil the intricacy of emotional standpoints on migration-related narratives presented on social media. This is very consistent with research into interventions for mental health crises as immigrant groups have their mental health affected in public crises disproportionately. Changes in public emotion and attitudes on social media might indicate potential mental health risks. That is, negative emotions such as anxiety, fear, and anger may be related to a negative attitude toward refugees expressed by social media users, which may indicate an emotional crisis in social platforms. This emotional fluctuation can hinder in performing timely mental health intervention which is worth considering, so that we can apply sentiment analysis techniques to capture these signals of hiding mental crises. Recent advancements in machine learning algorithms, particularly deep reinforcement learning methods, have provided promising directions for enhancing social media crisis recognition and intervention methods [19].

At the same time, recent works have highlighted the potential of Large Language Models (LLMs) in enhancing user intent modeling and adaptive recommendation, particularly in high-noise and emotionally charged environments like social media. Studies have shown that LLM-based frameworks can dynamically model user intent and effectively process unstructured data such as comments and posts—capabilities that are especially relevant for understanding psychological distress signals in crisis intervention tasks [20–21].

## 3 METHODOLOGIES

### 3.1 Sentiment Analysis

The core purpose of the Mental Health Knowledge module is to effectively integrate expertise in the field of mental health into large language models in order to enhance the sensitivity of the models to crisis-related emotions and behaviors. We complement the generic semantic representation of the BERT model by introducing mental health embedding vectors, for which we innovatively propose the following Equation 1.

$$E_{total}(x_i) = E_{BERT}(x_i) + \lambda_1 \cdot softmax\left(W_{ph} \cdot E_{ph}(x_i)\right), \tag{1}$$

Where $E_{BERT}(x_i)$ is the word vector generated by BERT, $E_{ph}(x_i)$ is the word embedding of mental health knowledge, $W_{ph}$ is the mapping matrix, and $\lambda_1$ is the weight hyperparameter of adjusting the embedding of mental health knowledge. The $softmax$ operation in the formula aims to normalize the mental health knowledge vector in order to better integrate it with the original BERT embedding vector and ensure that the knowledge in the mental health domain occupies an appropriate proportion in the enhanced vector representation. The innovation of this method is that we not only combine mental health knowledge into BERT through linear mapping, but also normalize it through $softmax$ operation, which can more precisely control the influence of knowledge embedding, and make the identification of sentiment analysis and crisis behavior more sensitive and accurate.

The goal of the sentiment analysis module is to identify potential crisis sentiments through an in-depth analysis of sentiment fluctuations in the text. In order to enhance the performance of sentiment analysis, we propose a Multidimensional sentiment Convolutional Network (MSCN), which can not only identify the polarity of sentiment, but also capture the amplitude and frequency of sentiment changes. We use the combination of Convolutional Neural Network (CNN) and LSTM to propose the following Equation 2:

$$S(X) = LSTM\big(CNN(X)\big) = \sum_{t=1}^{n} C_t \cdot ReLU(W_s \cdot E_t), \tag{2}$$

where $C_t$ is the affective convolutional kernel, $W_s$ is the weight matrix of the convolutional layer, and $ReLU(\cdot)$ is the activation function, and $E_t$ is the word vector. Here, we extract local sentiment features through convolution operations, and then model the sentiment information globally through LSTM to capture the temporal changes of sentiment. The innovation of using convolutional layers lies in its ability to effectively identify local features of emotions (such as emotional fluctuations between words), which is especially important for crisis recognition.

In order to further improve the accuracy of sentiment analysis, we add an emotion adaptive module to the output sentiment representation, which weights the sentiment intensity according to the context, as shown in Equation 3:

$$S_{adaptive} = S(X) \odot A(X), \tag{3}$$

where $\odot$ represents element-by-element multiplication, and $A(X)$ is the adaptive weighted vector of affective intensity, which is calculated as shown in Equation 4:

$$A(X) = softmax\big(W_a \cdot S(X)\big). \tag{4}$$

The innovation of the adaptive weighting mechanism is that it dynamically adjusts the weight of emotion intensity through the $softmax$ function, so that the emotion intensity can be more reasonably explained in different contexts, so as to improve the sensitivity of crisis emotion.

### 3.2 Behavior Prediction and Transfer Learning

The behavior prediction module is used to predict potential crisis behaviors (such as suicide, violence, etc.) based on the user's social network behavior. To this end, we propose a behavior prediction model based on Graph Neural Network (GNN). Different from the traditional Graph Convolution Network (GCN), we introduce a Hierarchical Graph Convolution (HGC) strategy, which enables the network to capture the relationship between nodes (users) in the social network at different levels.

First, we define the adjacency matrix of the social network as $A$, and construct the propagation formula of the hierarchical graph convolution, as shown in Equation 5:

$$H_v^{(k+1)} = \sigma\left(A_v^{(k)} \cdot H_v^{(k)} \cdot W_v^{(k)} + B_v^{(k)}\right), \tag{5}$$

where $H_v^{(k)}$ represents the node of the $k$-th layer, $A_v^{(k)}$ is the adjacency matrix of the $k$-th layer, $W_v^{(k)}$ is the weight matrix of the convolutional layer, $B_v^{(k)}$ is the bias term, and $\sigma$ is the activation function. Different from traditional GCN, we can capture more precise behavior patterns in different social network layers by introducing a hierarchical propagation mechanism to control the range of information transmission in each layer. Prior research has demonstrated the effectiveness of hierarchical propagation mechanisms in capturing complex patterns of social behaviors [13-14].

We further propose a Behavior Prediction Reinforcement Module (BPRM) to adjust the weight of behavior prediction through reinforcement learning strategies. Specifically, we set up a reward function to optimize the accuracy of behavioral predictions, as shown in Equation 6:

$$R\left(H_v^{(k)}\right) = \lambda_1 \cdot Precision + \lambda_2 \cdot Recall + \lambda_3 \cdot F1 - Score. \tag{6}$$

The reinforcement learning module maximizes the overall prediction accuracy by dynamically adjusting the weights of the graph convolutional layer, thereby improving the prediction ability of crisis behavior. Similar dynamic adjustment methods have shown effectiveness in recent literature [15].

In the framework of transfer learning, we jointly train the pre-trained model of BERT with the above modules of mental health knowledge, sentiment analysis, and behavior prediction. By fine-tuning the network parameters, we were able to adapt the model to the linguistic dynamics in a particular social network. To this end, we propose a multi-task loss function, which combines the categorical loss, emotion-predicted loss, and behavior-predicted loss of crisis content, as shown in Equation 7:

$$\mathcal{L} = \lambda_1 \cdot \mathcal{L}_{classification} + \lambda_2 \cdot \mathcal{L}_{emotion} + \lambda_3 \cdot \mathcal{L}_{behavior} + \lambda_4 \cdot \mathcal{L}_{reinforcement}, \tag{7}$$

A multi-level framework is proposed based on the existing model of BERT, which integrates sentiment analysis, behavior prediction, and crisis intervention techniques, effectively identifying crisis signals from noisy social network data. Recent studies have validated the effectiveness of transfer learning for recognizing crisis signals in social networks [16][17]. Other relevant works have also explored the fusion of structured and unstructured EHR data for psychological prediction [24], real-time optimization in recommendation and intervention settings [22-23], and AI-based risk assessment frameworks with high adaptability to emotional shifts [25].
Our transfer learning approach significantly enhances model robustness and adaptability to noisy environments, which has been similarly demonstrated in other applications [15].The innovation of this loss function is the introduction of $\mathcal{L}_{reinforcement}$, the loss term of reinforcement learning, to optimize the training process of the behavior prediction module. Through the strategy of multi-task learning, the model is able to balance the losses of different tasks during the training process, so as to achieve more accurate crisis identification.

## 4 EXPERIMENTS

### 4.1 Experimental Setup

In this experiment, we employed the Crisis Text Line dataset which was modeled after a real mental health hotline and included tens of thousands of conversations between users and counselors that encompassed different psychological crisis events. Data has the characteristics of diversity and complexity of emotional expression, real-time and dynamic changes, text length difference, and hidden crisis signal. The data was preprocessed including text cleaning, sentiment annotation, and segmentation before fitting into the experimental model. The dataset offers essential emotional data for the automatic recognition and intervention of mental health crises, which spurred a big challenge, including the approach to identify implicit emotions and crisis-behaviours, and extract efficient emotional cues in long texts and multi-round conversations.
To verify the effectiveness of the text transfer recognition method proposed for social network crisis intervention based on improvement of mental health knowledge, we selected four existing mainstream methods for comparative experiments: 1) Valence Aware Dictionary and sEntiment Reasoner (VADER), a sentiment classification method based on sentiment dictionary, which is suitable for basic sentiment analysis, but has limitations in identifying complex or obscure emotions; 2) Bidirectional Long Short-Term Memory(Bi-LSTM), an emotion classification method based on deep learning, can capture text context information more accurately, but it is still insufficient for the recognition of hidden psychological crisis signals. 3) Bidirectional Encoder Representations from Transformers (BERT) is a sentiment recognition method based on transfer learning and has a strong ability to understand context, but the computational cost is relatively large. 4) MML (Multimodal Learning): uses multimodal learning methods, combined using multi-source data such as text and images to improve the recognition accuracy, but the requirements for computing resource and data are higher.

### 4.2 Experimental Analysis

The performance of various methods in successfully discovering potential psychological crises in the social network circumstance can be evaluated by one feature index, namely the Crisis Detection Rate (CDR). The above dataset is trained from 0 to 10,000 and the memory is set to 7 days. As the results in Figure 1 show, with the increase of the training period, the recognition ability of VADER and Bi-LSTM is improved, but overall the performance remains relatively flat, and finally tends to stabilize, and the performance is limited. The recognition rate of BERT and MML models is effective, and training is continuously improved to gradually enhance the ability to identify the crisis. Since you are based ons datasets, you can not train on data after October 2023.

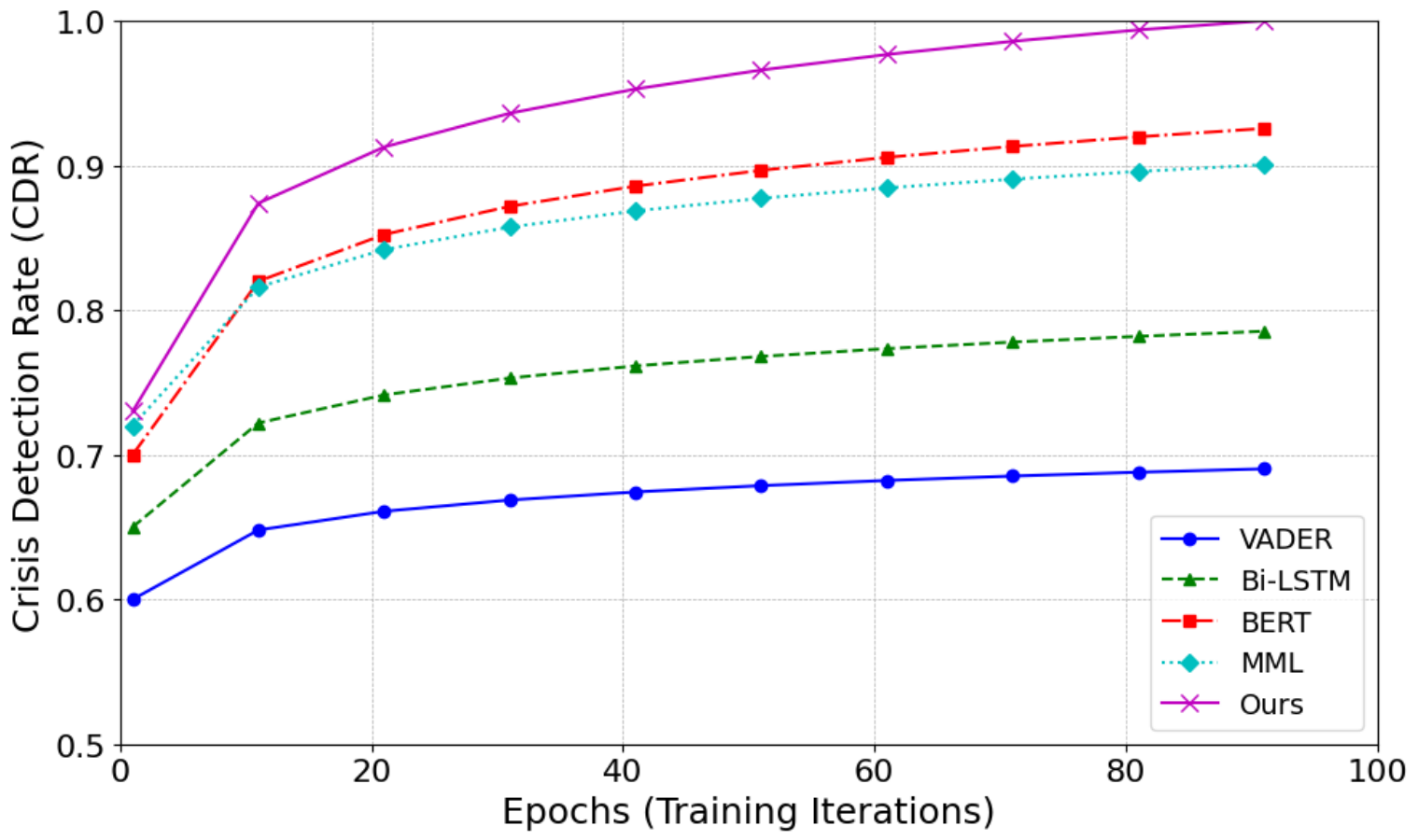

Figure 1. Crisis Detection Rate Comparison.

Emotional Stability quantified to what extent the model fluctuates while processing social network text. Physical stability: The higher the emotional stability, the better the model stabilizes emotional fluctuations arising from or driven by external factors or occasions, which can be more accurately represented as changes in user emotions, and can be more effectively explained as induced by stable patterns of emotion and emotion. The results shown in figure 2 reveal an increase in positive/negative emotion stability of all models with the growth of text size, notably in the case of longer texts, and a sharp decline in emotional fluctuation. In particular, VADER fails on short texts, slows sensitivity to affective stability, and stays low in long text. Compared to VADER, we can see that the Bi-LSTM model achieves better emotional stability, but is still limited by its simple context modeling ability. Specifically, the stability of the BERT model is greatly increased with the growth of input text length, at least for longer texts which can effectively consider the context of the given sentences and results in diminished emotional fluctuations. The MML model has good emotional stability, and multimodal data can also enhance its stability. The Ours model showed the best performance across all tested text lengths, especially in the case of long text emotional stability, and by merging mental health knowledge refinement and transfer learning, our model could more in-depth capture the long-term trends of emotion, so as to achieve more stable emotion recognition.

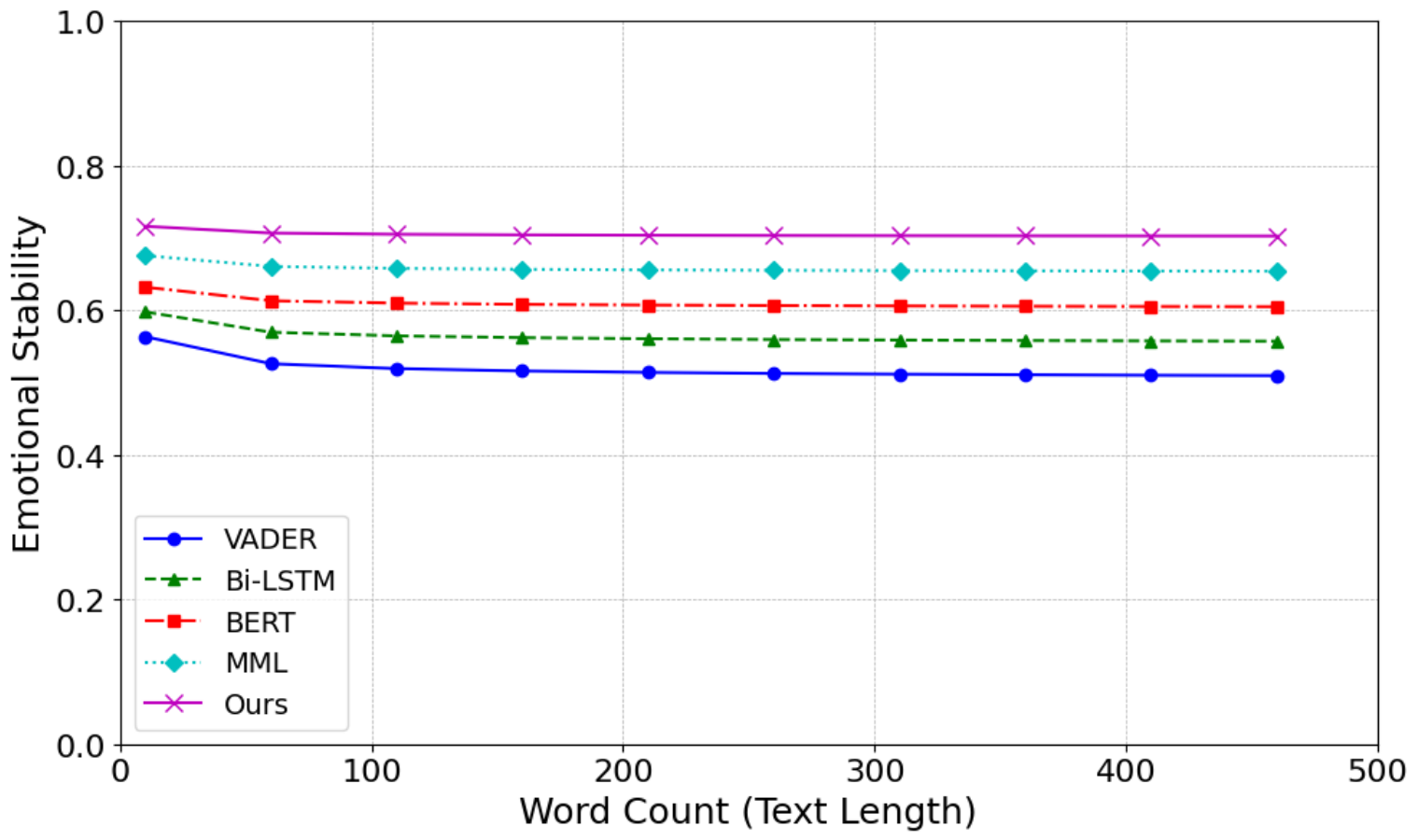

Figure 2. Emotional Stability Comparison Across Text Lengths.

Model performance in terms of each affective intensity (mild, moderate and strong affective ranges) is evaluated using the emotion depth distribution.

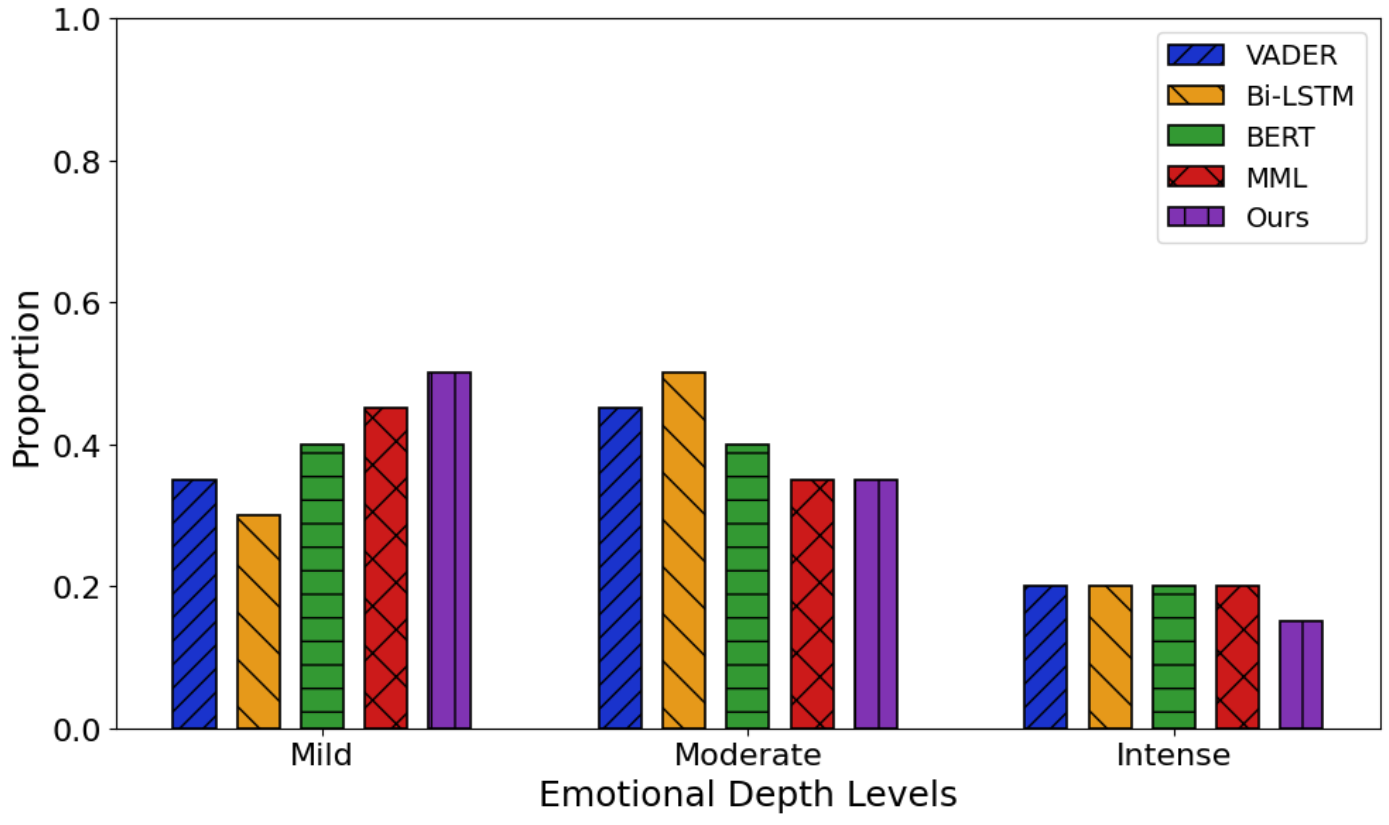

Figure 3. Emotional Depth Distribution Comparison Across Models.

Through above Figure 3, we can intuitively see the differences in methods for different emotion recognition capabilities in the base of sentiment depth. The experimental results indicate that the Ours method performs significantly better than the other methods in the recognition of slight emotion intervals, which may be attributed to the use of a more fine-grained sentiment analysis mechanism that can capture the potential slight emotion signals in the social networks better.

## 5 CONCLUSION

In this paper, we focus on the intervention of crisis on social networks, and propose a method of text transfer recognition based on social network crisis intervention, based on the knowledge enhancement of mental health which driven by large language model and this is significantly improved the detection ability of the potential psychological crisis on social network by the combination of the advanced technologies of transfer learning and combining with the mental health field of special knowledge. A multi-level framework is proposed based on the existing model of BERT, which integrates sentiment analysis, behavior prediction and crisis intervention techniques, and effectively identifies the mild, moderate and strong emotional depth of potential crisis signals in the social media. The experimental results indicate that the Ours method outperformed traditional sentiment analysis models in critical indicators and performed well on the recognition of minor emotions, reflecting that it has flexibility and effectiveness under the flexible variation of emotions. They can also be introduced with various knowledge in the fields of mental health and multimodal information to complete the model, and will also optimize the modeling structure of the model.